\documentclass[preprint,5p, times, twocolumn, authoryear]{elsarticle}

\usepackage{amsmath,amsfonts, bm}
\usepackage{array}
\usepackage{textcomp}
\usepackage{stfloats}
\usepackage{url}
\usepackage{verbatim}
\usepackage{graphicx}
\usepackage{cite}

\usepackage{enumitem}
\setlist[description]{font=\normalfont\itshape\space}
\usepackage{dsfont}
\usepackage[section]{placeins}
\usepackage{subcaption,graphicx}
\usepackage{mathrsfs}
\usepackage{float}
\usepackage[ruled,vlined,linesnumbered,lined,boxed,commentsnumbered]{algorithm2e}
\usepackage[colorlinks=true,citecolor=blue]{hyperref}
\usepackage[round]{natbib}

\usepackage{url}            
\usepackage{booktabs}       
\usepackage{thmtools, thm-restate}
\usepackage{amssymb}
\usepackage{nicefrac}       

\usepackage{placeins}
\usepackage{mathtools}
\usepackage{adjustbox}

\usepackage{multibib}
\usepackage{amsthm}
\usepackage{cleveref}

\newcommand{\E}[1]{\mathbb{E}\left[ #1 \right]} 
\newcommand{\V}[1]{{\mathbb{V}\left[#1 \right]}} 
\newcommand{\N}[2]{{\mathcal{N}\left(#1, #2\right)}}

\newtheoremstyle{named}{}{}{\itshape}{}{\bfseries}{}{.5em}{\thmnote{#3 }#1}
\theoremstyle{named}
\newtheorem*{namedtheorem}{Theorem}

\begin{document}

\begin{frontmatter}

\title{Optimal Training of Mean Variance Estimation Neural Networks}

\author{Laurens Sluijterman}
\address{Department of Mathematics, Radboud University, \\P.O. Box 9010-59, 6500 GL, Nijmegen, Netherlands}
\ead{L.Sluijterman@math.ru.nl}
\author{Eric Cator}
\address{Department of Mathematics, Radboud University}
\ead{e.cator@science.ru.nl}
\author{Tom Heskes}
\address{Institute for Computing and Information Sciences, Radboud University}
\ead{Tom.Heskes@ru.nl}

\begin{abstract}
This paper focusses on the optimal implementation of a Mean Variance Estimation network (MVE network) \citep{nix1994estimating}. This type of network is often used as a building block for uncertainty estimation methods in a regression setting, for instance Concrete dropout \citep{gal2017concrete} and Deep Ensembles \citep{lakshminarayanan2017simple}. Specifically, an MVE network assumes that the data is produced from a normal distribution with a mean function and variance function. The MVE network outputs a mean and variance estimate and optimizes the network parameters by minimizing the negative loglikelihood.

In our paper, we present two significant insights. Firstly, the convergence difficulties reported in recent work can be relatively easily prevented by following the simple yet often overlooked recommendation from the original authors that a warm-up period should be used. During this period, only the mean is optimized with a fixed variance. We demonstrate the effectiveness of this step through experimentation, highlighting that it should be standard practice. As a sidenote, we examine whether, after the warm-up, it is beneficial to fix the mean while optimizing the variance or to optimize both simultaneously. Here, we do not observe a substantial difference.  Secondly, we introduce a novel improvement of the MVE network: separate regularization of the  mean and the variance estimate. We demonstrate, both on toy examples and on a number of benchmark UCI regression data sets, that following the original recommendations and the novel separate regularization can lead to significant improvements. 
\end{abstract}
\end{frontmatter}

\section{Introduction}
\noindent 
Neural networks are gaining tremendous popularity both in regression and classification applications.  In a regression setting, the scope of this paper, neural networks are used for a wide range of tasks such as the prediction of wind power \citep{khosravi2014optimized}, bone strength \citep{shaikhina2017handling}, and floods \citep{chaudhary2022flood}.

Due to the deployment of neural networks in these safety-critical applications, uncertainty estimation has become increasingly important \citep{gal2016uncertainty}. The uncertainty in the prediction  can be roughly decomposed into two parts: epistemic or model uncertainty, the reducible uncertainty that captures the fact that we are unsure about our model, and aleatoric uncertainty, the irreducible uncertainty that arises from the inherent randomness of the data \citep{hullermeier2019aleatoric, abdar2021review}. In this paper, we refer to the latter as the variance of the noise, to avoid any confusion or philosophical discussions. The variance of the noise can be \textit{homoscedastic} if it is constant, or \textit{heteroscedastic} if it depends on the input $\bm{x}$.

There is a vast amount of research that studies the model uncertainty. Notable approaches include Bayesian neural networks \citep{mackay1992practical, neal2012bayesian}, dropout \citep{gal2016dropout, gal2017concrete}, and ensembling \citep{heskes1997practical}. Conversely, a lot less emphasis is often placed on the estimation of the variance of the noise. Monte-Carlo dropout, for example, simply uses a single homoscedastic hyperparameter. Some other methods, such as concrete dropout and the hugely popular Deep Ensembles \citep{lakshminarayanan2017simple}, use a Mean Variance Estimation (MVE) network \citep{nix1994estimating}.

\begin{figure*}[tb] 
\vskip-0.1in
\centering
\begin{minipage}[t]{0.49\textwidth}
	\includegraphics[width=0.8\textwidth]{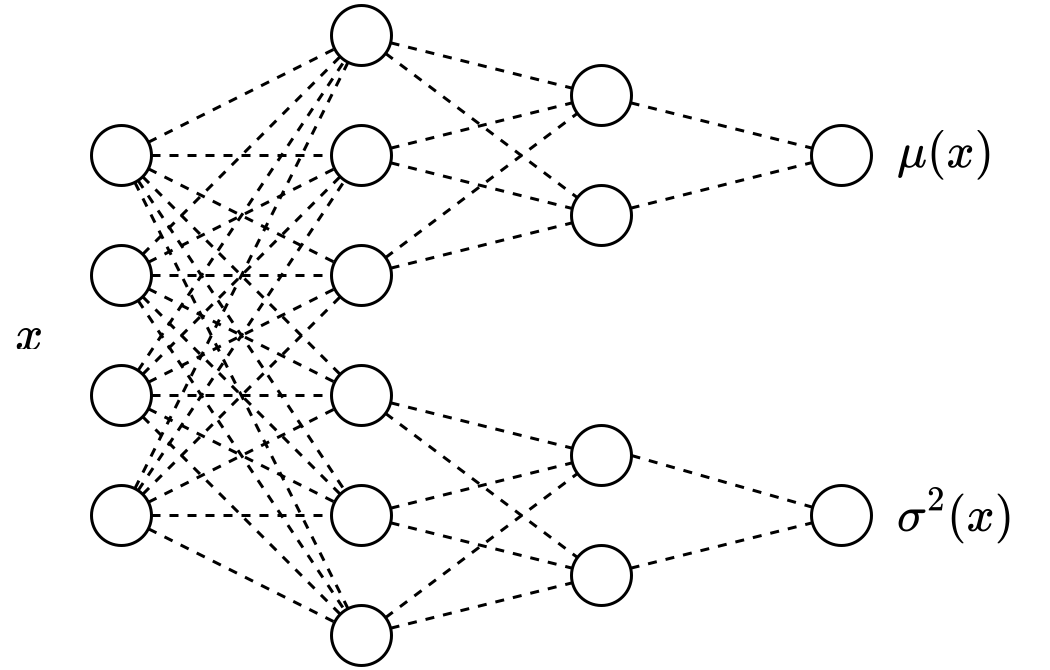}
	\caption{Original MVE architecture}
	\label{fig: originalMVE}
\end{minipage}
\begin{minipage}[t]{0.49\textwidth}
	\includegraphics[width=0.8\textwidth]{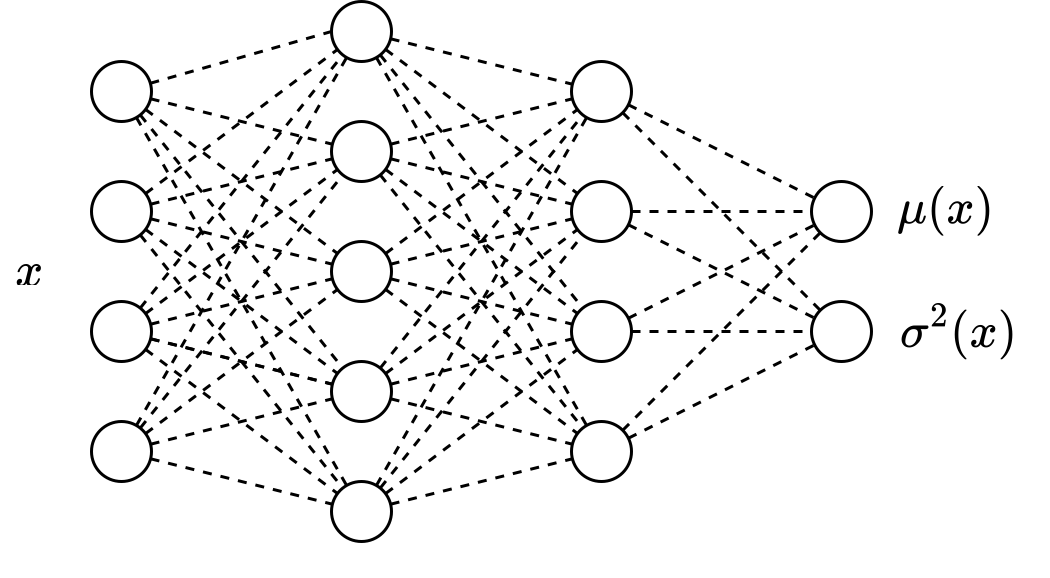}
	\caption{Modern MVE architecture}
	\label{fig: modernMVE}
\end{minipage}
\vskip -0.1in
\end{figure*}

An MVE network, see Figure \ref{fig: originalMVE}, works as follows. We assume that we have a data set consisting of $n$ pairs $(\bm{x}_{i}, y_{i})$, with $y_{i} \sim \N{\mu(\bm{x}_{i})}{\sigma^{2}(\bm{x}_{i})}$. An MVE network consists of two sub-networks that output a prediction for the mean, $\mu_{\theta}(\bm{x})$, and for the variance, $\sigma^{2}_{\theta}(\bm{x})$. These sub-networks only share the input layer and do not have any shared weights or biases. In order to enforce positivity of the variance, a transformation such as a softplus or an exponential is used. The network is trained by using the negative loglikelihood of a normal distribution as the loss function:
\[
\mathcal{L}(\theta) = \sum_{i=1}^{N} {1 \over 2}\log(\sigma^{2}_{\theta}(\bm{x}_{i})) + {1 \over 2}\frac{(y_{i} - \mu_{\theta}(\bm{x}_{i}))^{2}}{\sigma^{2}_{\theta}(\bm{x}_{i})}
\]

Since the MVE network is often used as the building block for complex uncertainty estimation methods, it is essential that it works well. Multiple authors have noted that the training of an MVE network can be unstable \citep{seitzer2021pitfalls, skafte2019reliable, takahashi2018student}. The main argument, elaborated on in the next section, is that the network will start focussing on areas where the network does well at the start of the training process while ignoring poorly fitted regions. However, \citet{nix1994estimating} already warned for the possibility of harmful overfitting of the variance and gave the solution:
\begin{center}
\textit{The training of an MVE network should start with a warm-up period where the variance is fixed and only the mean is optimized.}
\end{center}
Additionally, the variance is initialized at a constant value in order to make all data points contribute equally to the loss. \citet{nix1994estimating} did not demonstrate the importance of this warm-up period in the original paper. In this paper, we empirically demonstrate that using a warm-up period can greatly improve the performance of MVE networks and fixes the instability noted by other authors.

 A limited amount of research has investigated possible improvements of the MVE network \citep{seitzer2021pitfalls, skafte2019reliable}. Most improvements require a significant adaptation to the training procedure such as a different loss function or locally aware mini-batches. However, to the best of our knowledge, none have investigated our proposed easy-to-implement improvement:
\begin{center}
\textit{The mean and variance in an MVE network should be regularized separately.}
\end{center}	
Most modern methods \citep{jain2020maximizing, egele2021autodeuq, gal2017concrete, lakshminarayanan2017simple} appear to use the same regularization for both the mean and the variance. In fact, the current use of the MVE network often does not even easily allow for different regularizations. Typically, only a second output node is added to represent the variance, instead of an entire separate sub-network (see Figure \ref{fig: modernMVE}). As we will demonstrate in this paper, separate regularization can be very beneficial to the predictive results.

\noindent \textit{Contributions:}
\begin{itemize}
	\item We provide experimental results that demonstrate the importance of a warm-up period as suggested by \citet{nix1994estimating}.
	\item We explore the benefits of updating the mean and variance simultaneously after the warm-up versus solely learning the variance while keeping the mean fixed. 
	\item We offer a theoretical justification for why distinct regularization of the mean and variance is essential for an MVE network. We back up our claims with real-world evidence, demonstrating how this approach can lead to significant enhancements.
\end{itemize}

\noindent \textit{Organisation:} 

This paper consists of 5 sections, this introduction being the first. In Section \ref{trainingdifficulties}, we go through the problems with MVE networks that have recently been reported in the literature. In the same section, we show that these problems can be resolved by following the recommendation of using a warm-up period. We also provide additional theoretical motivation in favor of updating both the mean and the variance after the warm-up as opposed to keeping the mean fixed and only learning the variance. Section \ref{separaterularization} explains, both intuitively and using classical theory, why we expect to need different amounts of regularization for the mean and the variance estimates. Both the effect of the warm-up and of separate regularization are experimentally examined in Section \ref{Experimental}. The final section summarizes the results, gives a list of recommendations when training an MVE network, and provides possible avenues for future work.

All the code used in the experiments of this paper can be found at \url{https://github.com/LaurensSluyterman/Mean_Variance_Estimation}.

\section{Difficulties with training MVE networks} \label{trainingdifficulties}
\noindent It is known that the training of an MVE network can be unstable \citep{seitzer2021pitfalls, skafte2019reliable, takahashi2018student}. The main argument is that the network may fail to learn the mean function for regions where it initially has a large error. In these regions, the variance estimate will increase, which implies that the residual does not contribute to the loss as much. The network will start to focus more on regions where it is performing well, while increasingly ignoring poorly fit regions.

\begin{figure}[tb]
\centering
\vskip 0in
\includegraphics[width=0.45\textwidth]{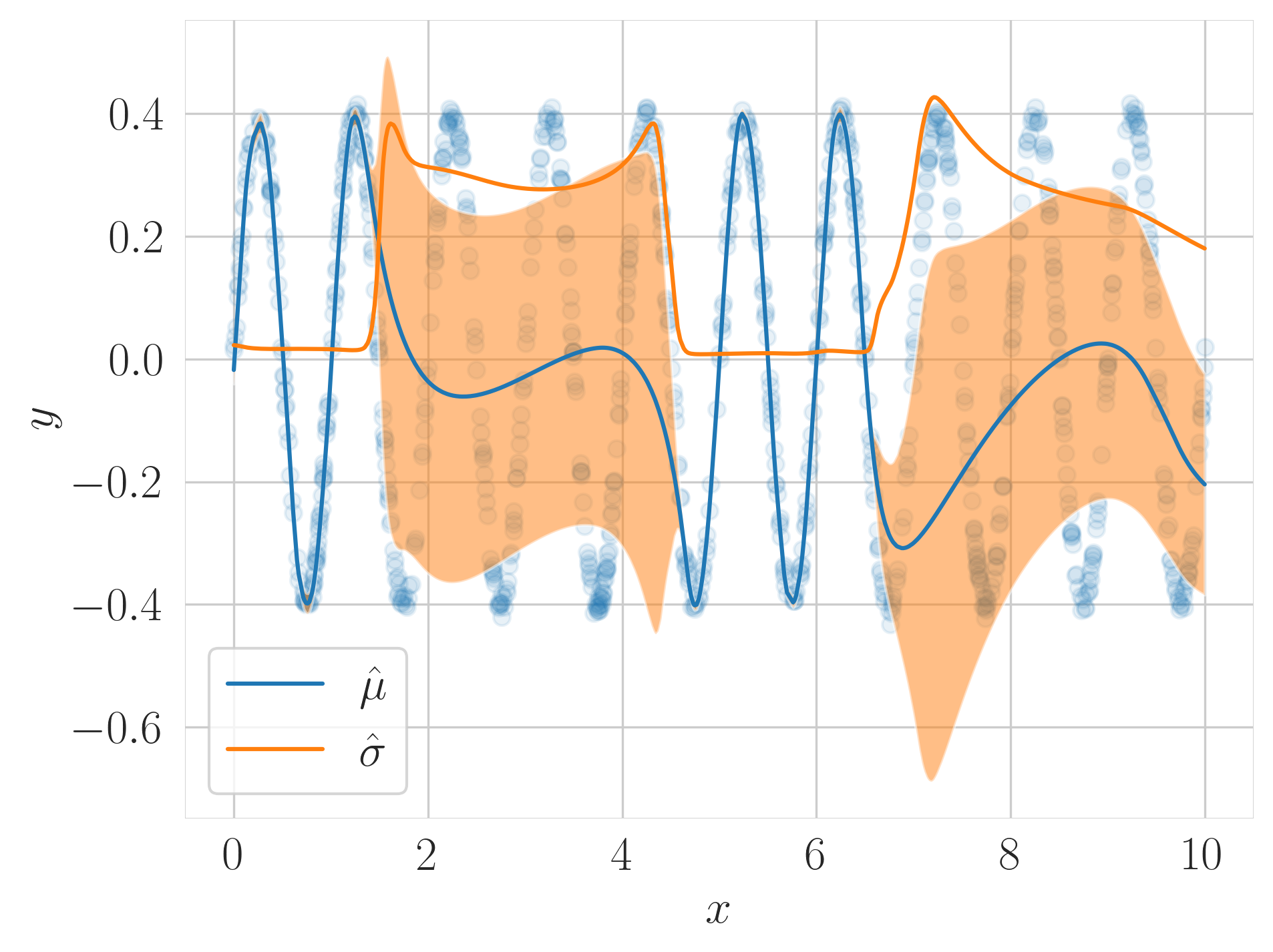}
\caption{An example of an MVE network that fails to learn the mean function when simultaneously updating the mean and the variance. The orange area gives plus or minus a single standard deviation.}
\label{fig: sin}	
\centering
\vskip 0in
\end{figure}

To illustrate what can happen, we reproduced an experiment from \citet{seitzer2021pitfalls}. We sampled 1000 covariates, $x_{i}$, uniformly between 0 and 10, and subsequently sampled the targets, $y_{i}$, from a $\N{0.4\sin(2\pi x_{i})}{0.01^2}$ distribution. Figure \ref{fig: sin} shows that the MVE network is unable to properly learn the mean function. Increasing training time does not solve this. A network with a similar architecture that was trained using the mean-squared-error loss \textit{was} able to learn the mean function well.

\begin{figure}[h]
\centering

\centering
\includegraphics[width=0.45\textwidth]{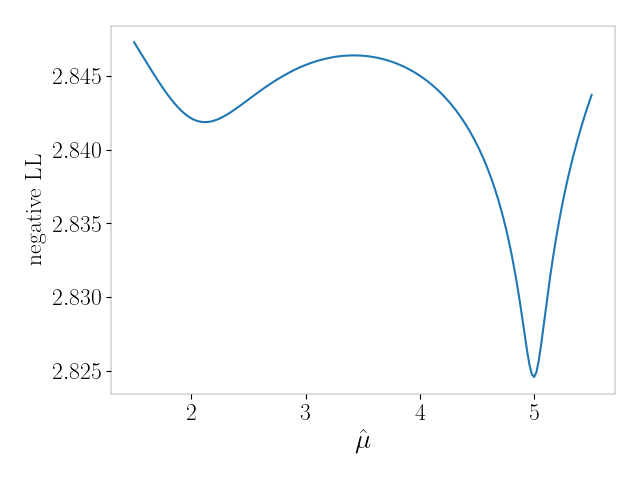}
\caption{A simple example of local minima in the negative loglikelihood. The data consist of two parts: 100 data points from a $\N{2}{0.5^2}$ distribution and 100 data points from a $\N{5}{0.1^{2}}$ distribution. The graphs shows negative loglikelihood as a function of $\hat{\mu}$ where we take the optimal variance estimates for each value of $\hat{\mu}$.}
\label{fig: localminima}
\centering
\end{figure}

We provide a second explanation for this behaviour by noting that the loss landscape is likely to have many local minima. We already encounter this in a very simple example. Suppose we have a data set consisting of two parts: 100 data points from a $\N{2}{0.5^2}$ distribution and 100 data points from a $\N{5}{0.1^{2}}$ distribution. For each part, we are allowed to pick a separate variance estimate, $\hat{\sigma}_{1}^{2}$ and $\hat{\sigma}_{2}^{2}$ but we can only pick a single estimate for the mean. In this situation, there are two local minima of the negative loglikelihood (see Figure \ref{fig: localminima}): we can set $\hat{\mu}$ to approximately 2 with a small $\hat{\sigma}_{1}^{2}$ and large $\hat{\sigma}_{2}^{2}$ or set $\hat{\mu}$ to 5 with a large $\hat{\sigma}_{1}^{2}$ and small $\hat{\sigma}_{2}^{2}$. 

While this simplified setting is of course not a realistic representation of a neural network, it does illustrate that there can easily be many local minima when dealing with complex functions for the mean and the variance. When we start from a random estimate for the mean, it is therefore not unlikely to end up in a bad local minimum.

\subsection{The solution: Warm-up}
\noindent The original authors advised to use a warm-up, which indeed alleviates most problems. After initialization, the variance is fixed at a constant value and the mean estimate is learned. In a second phase, the mean and variance are updated simultaneously. 

\begin{figure}[h]
\centering
\centering
\vskip -0 in
\includegraphics[width=0.45\textwidth]{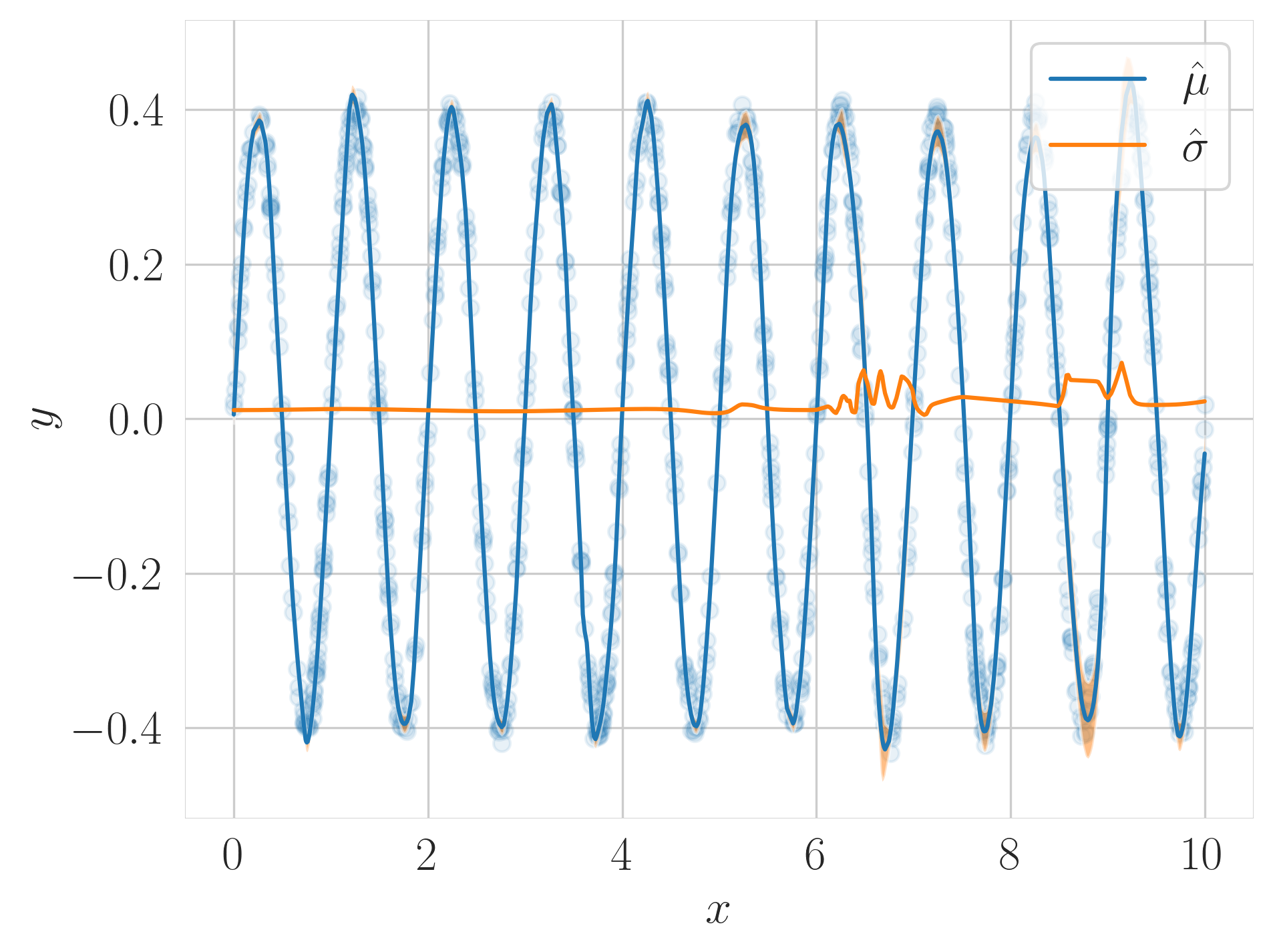}
\caption{By using a warm-up where only the mean is updated, the MVE network is able to learn the mean function well. In this example, the variance appears to be overfitting slightly.}
\label{fig: sinwarmup}	
\centering
\vskip -0 in
\end{figure}

We can motivate why a warm-up is beneficial, both from the loss-contribution perspective and from the local minima perspective. From the loss-contribution perspective, when keeping the variance fixed during the warm-up, we do not have the problem that regions that are predicted poorly initially fail to learn. Because the variance estimate at initialization is constant, all residuals contribute to the loss equally. From the loss-landscape perspective, we are less likely to end up in a bad local minima if we start from a sensible mean function. Figure \ref{fig: sinwarmup} shows that adding a warm-up period indeed solves the convergence problem that we previously had in the sine example in Figure \ref{fig: sin}.

\subsection{What to do after the warm-up?} \label{afterwarmup}
\noindent After the warm-up period, we could either update the variance while keeping the mean estimate fixed or update both simultaneously. In the original MVE paper, the authors argue that simultaneously estimating the mean and the variance is also advantageous for the estimate of the mean. The reasoning is that the model will focus its resources on low noise regions, leading to a more stable estimator. 

From a general theoretical perspective, there are clear advantages to optimizing the full likelihood. The resulting maximum-likelihood-estimate is consistent and asymptotically efficient (see any standard textbook on statistics, for instance \citet[chapter~7]{degroot1986Probability}). No other consistent estimator can asymptotically have a lower variance. For a linear model, a similar result even holds for the non-asymptotic regime: Taking the variance of the noise into account leads to an estimator with lower variance. We provide additional details on these statements in \ref{advantages}.

This lower variance in turn results in improved metrics such as RMSE. We demonstrate this both for the general case and for a linear model in  \ref{generalcase} and \ref{linearmodeldetails1}. These theoretical results illustrate that, besides the obvious benefit of having an estimate of the variance, it is also beneficial for the mean estimate to take the variance into account. Even if we measure performance on unseen data with the mean-squared error, there are valid arguments to take the variance of the residuals into account when estimating the mean.

In summary, focussing on low noise regions is beneficial. However, the estimate of the noise strongly depends on the quality of the mean predictor. If the mean predictor is bad, the estimation will not focus on low noise regions but on high accuracy regions, which can be very detrimental. We therefore need a warm-up period, after which classical theory would suggest that estimating the mean and variance simultaneously has advantages. In Section \ref{Experimental}, we test whether estimating the mean and variance simultaneously is indeed beneficial for the mean estimate.

\section{The need for separate regularization} \label{separaterularization}
\noindent In this section, we give a theoretical motivation for the need for a different regularisation of the parts of the network that give the mean and variance estimate. The amount of regularization that is needed when estimating a function depends on the smoothness and there is no reason to assume that the mean function and the variance function are equally smooth. If one function is much smoother than the other, we should not regularize them the same way. For instance, in the case of homoscedastic additive noise, the variance function is a constant function, whereas the mean function is likely not. 

We can make this argument explicit for a classical linear model. We do this by considering two linear models that most closely resemble the scenario of an MVE network. The first model will estimate the mean while knowing the variance and the second model will estimate the log of the variance\footnote{An MVE network often uses an exponential transformation in the output of the variance neuron to ensure positivity. The network then learns the log of the variance.} while knowing the mean. Both models will in general have a different optimal regularization constant. All derivations for the properties of linear models used in this paper can be found in \citet{van2015lecture}.

\begin{figure*}[tb] 
\centering
\vskip-0in
\subcaptionbox{High regularization: 0.1}{\includegraphics[width=0.24\linewidth]{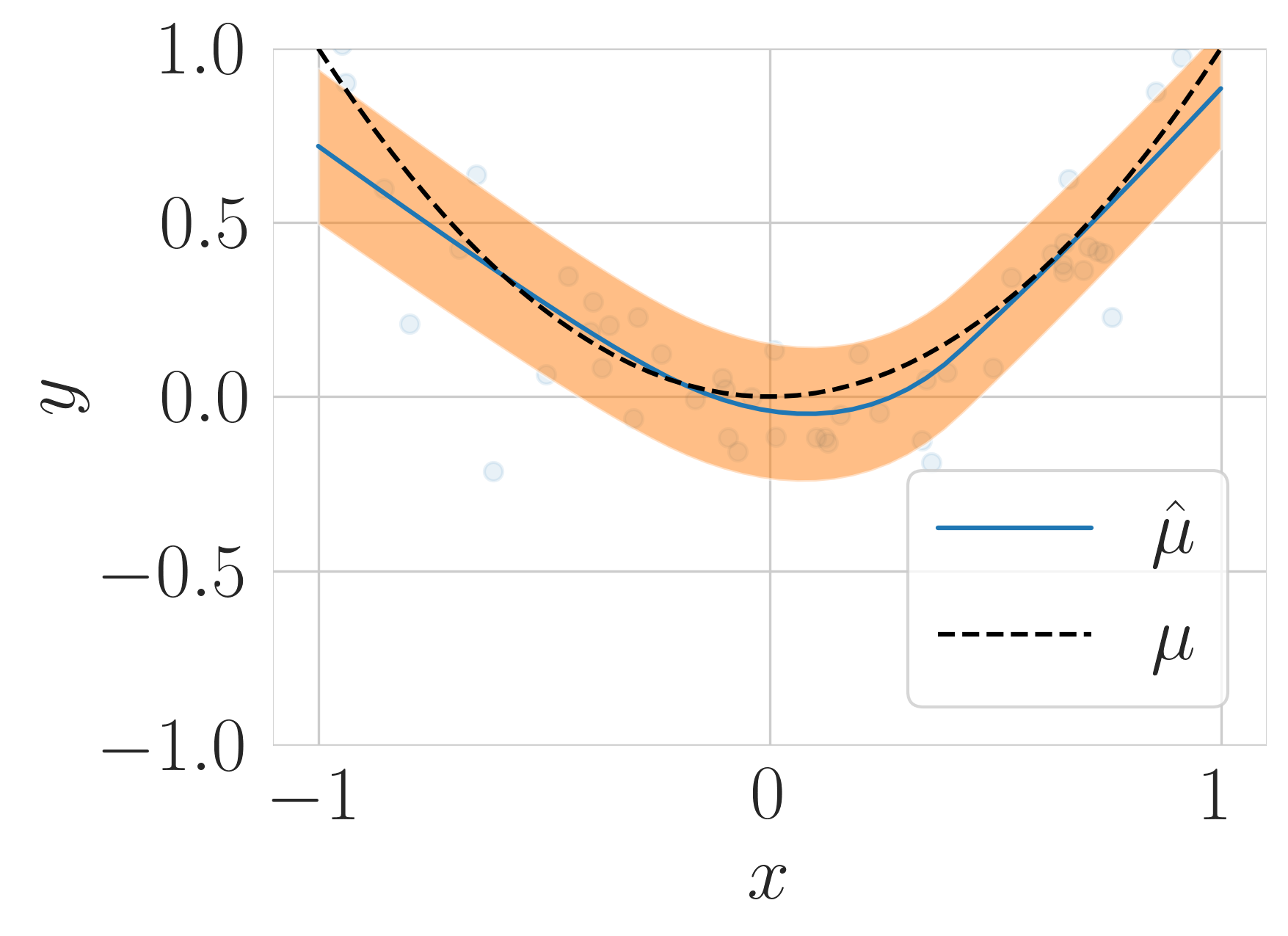}}
\subcaptionbox{Medium regularization: 0.04}{\includegraphics[width=0.24\linewidth]{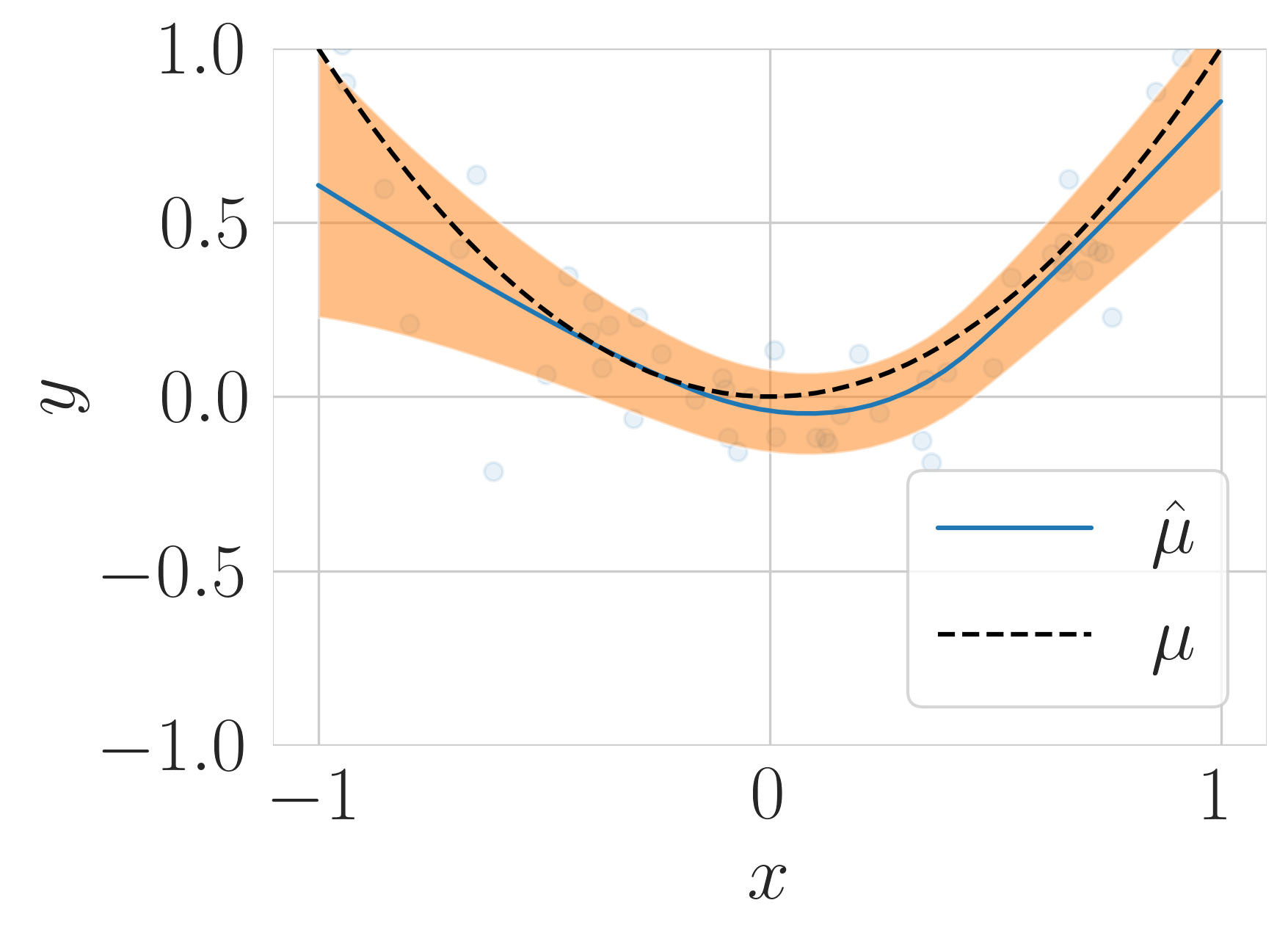}}
\subcaptionbox{Low regularization: 0.01}{\includegraphics[width=0.24\linewidth]{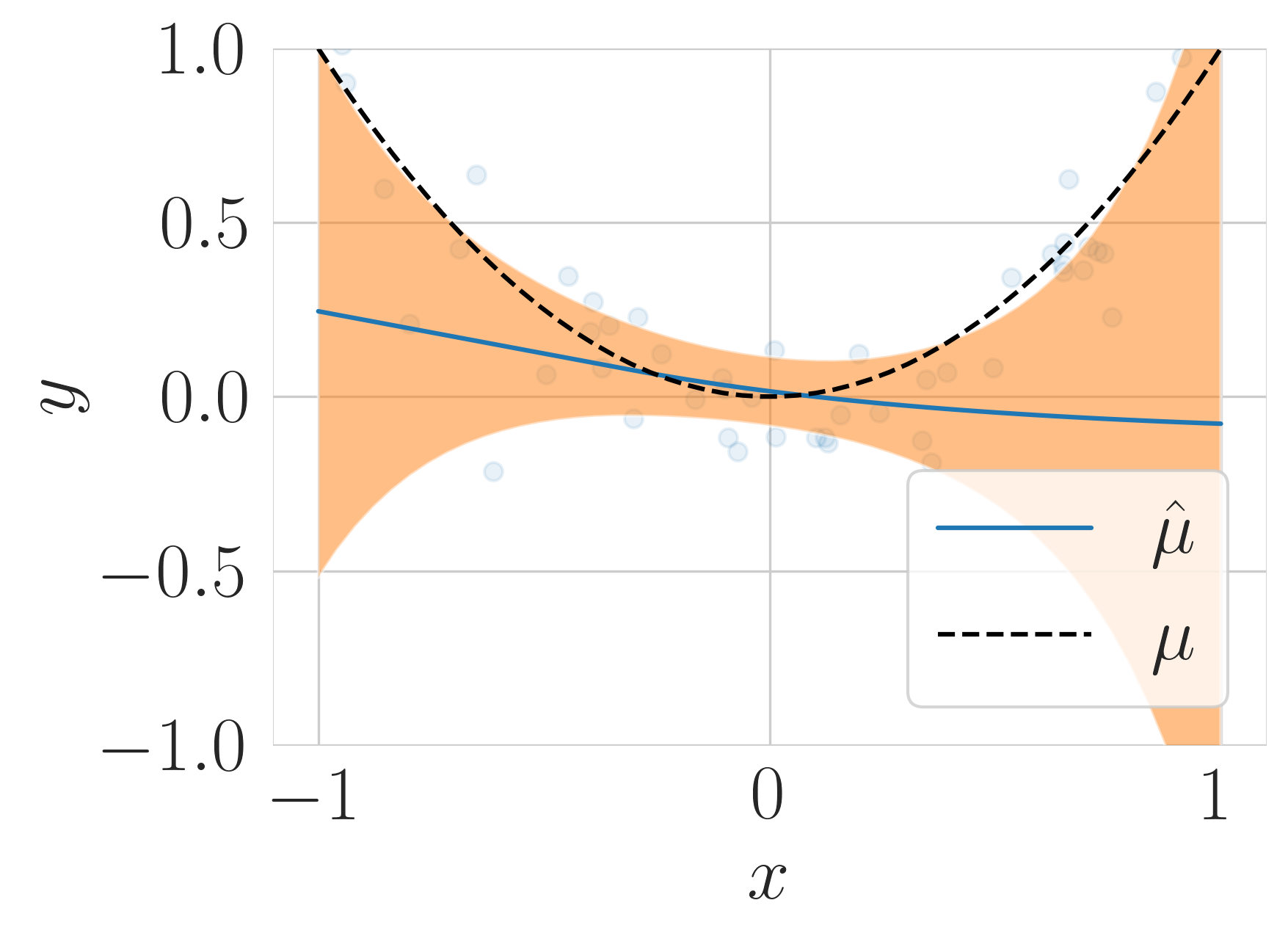}}
\subcaptionbox{No regularization: 0}{\includegraphics[width=0.24\linewidth]{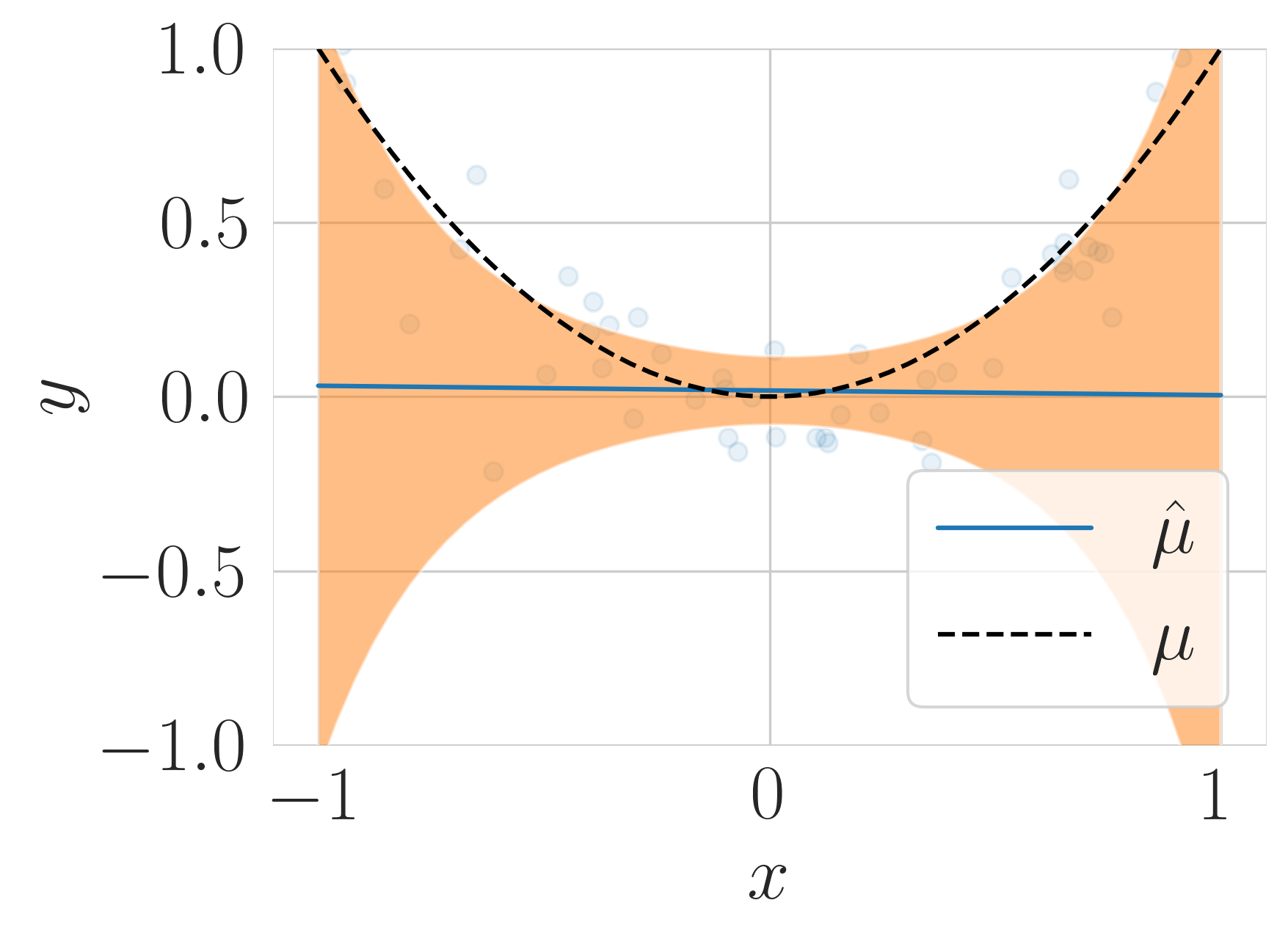}}
\caption{The effect of the different amounts of regularization of the variance. In all four figures, the mean has the same regularization constant of 0.1. The regularization of the variance is given in the subcaptions. The orange area gives plus or minus a single standard deviation.}
\vskip -0.1in
\label{fig: harmful_variance}
\end{figure*}
We acknowledge that a neural network is much more intricate than a linear model. However, since even for a simple linear model it is essential to have different regularization, the same applies to more complex models like neural networks that have a linear model as a special case. The empirical results that follow later corroborate this.

\subsection{Scenario 1: Estimating the mean with a known variance}

\noindent We consider a linear model with unknown homoscedastic noise. Specifically, we assume that we have a data set consisting of $n$ data points $(\bm{x}_{i}, y_{i})$, with $\bm{x}_{i} \in \mathbb{R}^{p}$ and $y \in \mathbb{R}$. With $X$, we denote the $n 
\times p$ design matrix which has the $n$ covariate vectors $\bm{x}_{i}$ as rows. With $Y$, we denote the $n\times 1$ vector containing the observations $y_{i}$. We assume $X$ to be of full rank and consider the following model:
\begin{equation}
Y = X\beta + U, \quad U \sim \N{0}{\sigma^{2} I_{n}}.
\label{eq: linearmodel}
\end{equation}
The goal is to find the $\beta\in \mathbb{R}^{p}$ that minimizes the total squared error plus a regularizaiton term:
\[
\sum_{i=1}^{n}(y_{i} - \bm{x}^{T}_{i}\beta)^{2} + \lambda \sum_{j=1}^{p}\beta_{j}^{2}.
\]
Different values of $\lambda$ result in different estimators $\hat{\beta}(\lambda)$. In \ref{linearmodeldetails2}, we show that optimal regularization constant, $\lambda^{\ast}$, satisfies

\[
\lambda^{\ast} \propto p (\beta^{T}\beta)^{-1},
\]
where we defined optimal as the $\lambda$ for which

\[
\text{MSE}(\hat{\beta}(\lambda)):= \E{|\beta - \hat{\beta}(\lambda)|^{2}}
\]
is minimal.

\subsection{Scenario 2: Estimating the log-variance with a known mean}
\noindent Next, we examine a linear model that estimates the logarithm of the variance. We again have $n$ datapoints $(\bm{x}_{i}, y_{i})$ and we assume the log of the variance to be a linear function of the covariates:
\[
y_{i} = \mu_{i} + \epsilon, \quad \epsilon \sim \N{0}{e^{\bm{x}_{i}^{T}\tilde{\beta}}}
\]
We use the same covariates and for the targets we define:
\[
z_{i} := \log((y_{i} - \mu)^{2})) - C, \quad \text{with} \quad C = \psi(1/2) +\log(2),
\]
where $\psi$ is the digamma function. This somewhat technical choice for $C$ is made such that
\[
z_{i} = \log(\sigma^{2}(\bm{x}_{i})) + \tilde{\epsilon}, 
\]
where $\tilde{\epsilon}$ has expectation zero and a constant variance. The details can be found in \ref{linearmodeldetails2}. In the same appendix we repeat the same procedure, i.e. minimizing the mean-squared-error with a regularization term, and demonstrate that the optimal regularization constant, $\lambda^{\ast}$, satisfies
\[
\tilde{\lambda}^{\ast} \propto p (\tilde{\beta}^{T}\tilde{\beta})^{-1}.
\]
The conclusion is that for these two linear models, that most closely resemble the scenario of regularized neural networks that estimate the mean and log-variance, the optimal regularization constants rely on the true underlying parameters $\beta$ and $\tilde{\beta}$. Since there is no reason to assume that these are similar, there is also no reason to assume that the mean and variance should be similarly regularized.

\subsection{Separate regularization of the variance alleviates the variance-overfitting}
\noindent While the problem of ignoring initially poorly fit regions is still present, proper regularization of the variance can alleviate the harmful overfitting of the variance. To illustrate this effect, we trained 4 MVE networks, without a warm-up period, on a simple quadratic function with heteroscedastic noise. The $x$-values were sampled uniformly from $[-1,1]$ and the $y$-values were subsequently sampled from a $\N{x^{2}}{(0.1 + 0.2x^{2})^{2}}$ distribution. We used the original MVE architecture which has two sub-networks that estimate the mean and the variance. We used separate $l_2$-regularization constants for both sub-networks in order to be able to separately regularize the mean and the variance. We used the same mean regularization in all networks and gradually decreased the regularization of the variance. 

Figure \ref{fig: harmful_variance} demonstrates the effect of different amounts of regularization of the variance. When the variance is regularized too much, the network is unable to learn the heteroscedastic variance. This is problematic both because the resulting uncertainty estimates will be wrong, but also because we lose the beneficial effect on the mean that we discussed in the previous subsection. In the second subfigure, the network was able to correctly estimate both the mean and variance. When we decreased the regularization of the variance further, however, we see that the network simply increased the variance on the right side instead of learning the function. When we remove regularization of the variance all together, the network was completely unable to learn the mean function.

\begin{figure}[t]
\vskip -0in
\centering
\includegraphics[width=0.42\textwidth]{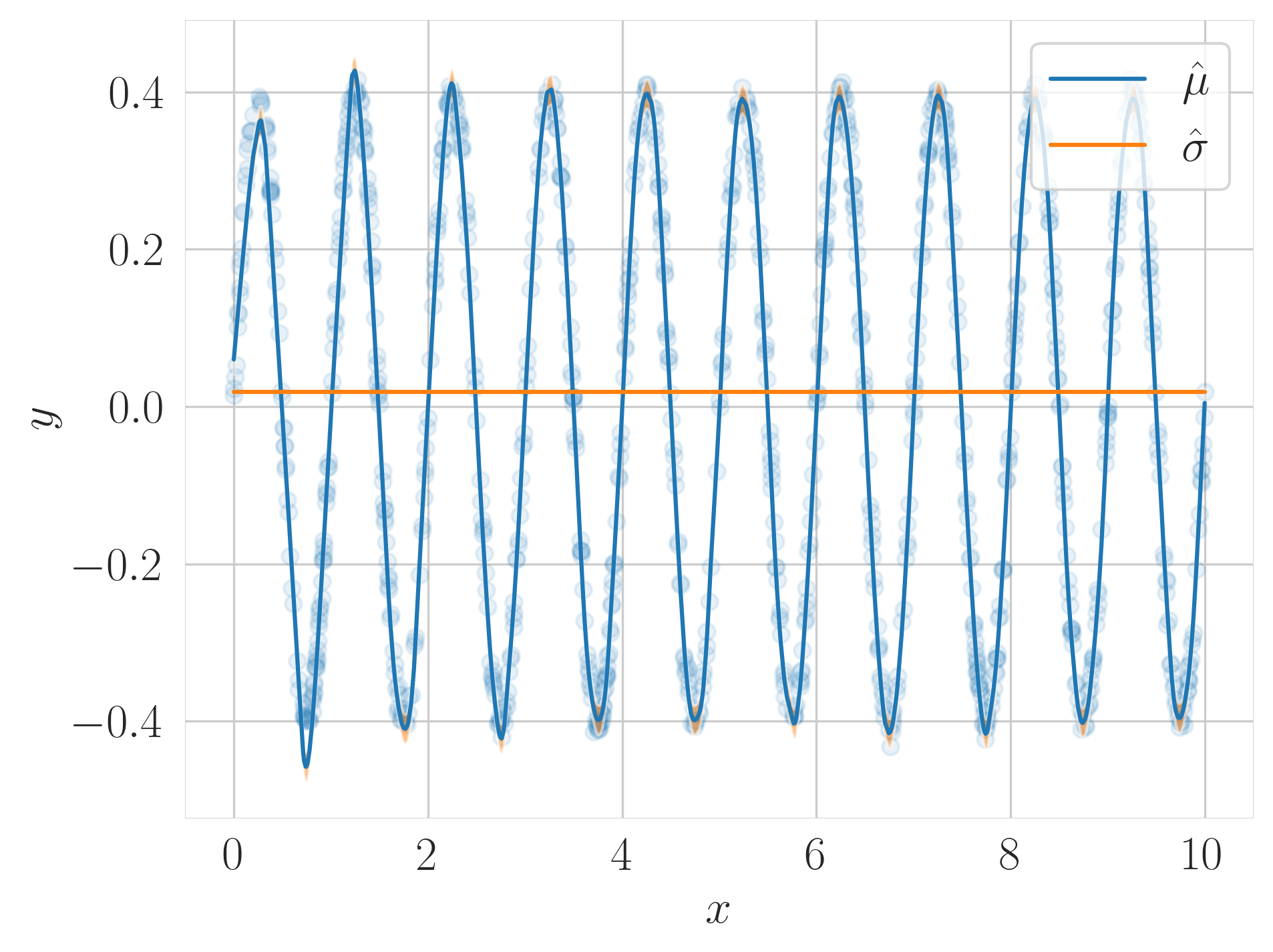}
\caption{By using a separate regularization constant for the variance, the MVE network is able to learn both the mean and the variance function well.}
\label{fig: sinseparate}	
\vskip -0.2in
\end{figure}

Additionally, we repeated the sine experiment while using a higher regularization constant for the variance than for the mean. In Figure \ref{fig: sinseparate}, we see that the MVE network is now able to learn the sine function well, even without a warm-up period. We were unable to achieve this when using the same regularization constant for both the mean and the variance.

\section{Experimental results} \label{Experimental}
\noindent In this section, we experimentally demonstrate the benefit of a warm-up period and separate regularization. In Subsection \ref{approaches}, we specify the three training strategies that we compare. Subsections \ref{datasets} and \ref{architecture} give details on the data sets, experimental procedure, and architectures that we use. Finally, the results are given and discussed in Subsection \ref{resultsanddiscussion}.

\subsection{Three approaches} \label{approaches}
\noindent We compare three different approaches:
\begin{enumerate}
	\item No Warm-up: This is the approach that is used in popular methods such as Concrete dropout and Deep Ensembles. The mean and the variance are optimized simultaneously.
	\item Warm-up: This is the approach recommended in the original paper. We first optimize the mean and then both the mean and the variance simultaneously.
	\item Warm-up fixed mean: We first optimize the mean and then optimize the variance while keeping the mean estimate fixed. We add this procedure to test if optimizing both the mean and the variance after the warm-up further improves the mean estimate. 
\end{enumerate}

\noindent For each approach, we consider two forms of $l_{2}$-regularization:
\begin{enumerate}
	\item Separate regularization: The part of the network that estimates the mean has a different regularization constant than the part of the network that estimates the variance.
	\item Equal regularization: Both parts of the network use the same regularization constant.
\end{enumerate}
\subsection{Data sets and experimental procedure} \label{datasets}
\noindent We compare the three approaches on a number of regression UCI benchmark data sets. These are the typical regression data sets that are used to evaluate neural network uncertainty estimation methods \citep{gal2016dropout, lakshminarayanan2017simple, hernandez2015probabilistic, pearce2018high}.

For each data set we use a 10-fold cross-validation and report the average loglikelihood and RMSE on the validation sets along with the standard errors. For each of the 10 splits, we use another 10-fold cross-validation to obtain the optimal regularization constants. The entire experimental procedure is given in Algorithm \ref{Procedure}.

\begin{algorithm}[t]
\textbf{Input:} Data set $(X, Y)$\;
Devide $(X, Y)$ in 10 distinct subsets, denoted $(X^{(i)}, Y^{i})$\;
\For{i \text{from} $1$ \text{to} $10$}{
  $X_{\text{train}} = \cup_{j \neq i}X^{(j)},$ $Y_{\text{train}} = \cup_{j \neq i}Y^{(j)},$ $X_{\text{val}} = X^{(i),}$ $Y_{\text{val}} = Y^{(i)}$\;
  Use 10-fold cross-validation (using only ($X_{\text{train}}, Y_{\text{train}}$)) to find the optimal regularization constants. This is done by choosing the constants for which the loglikelihood on the left-out sets is highest. The possible regularization constants are $[0.00001, 0.0001, 0.001, 0.01, 0.1]$\;
  Train a model using the optimal separate regularization constants\;
  Train a model using the optimal equal regularization constant\;
  Evaluate the loglikelihood and root-mean-squared error on the validation set\;}
\textbf{Return:} The average of the 10 loglikelihood and RMSE values along with the standard error.\;
\caption{Our experimental procedure. A 10 fold cross-validation is used to compare the different methods. In each fold, a second 10-fold cross-validation is used to obtain the optimal regularization constants. We use the same splits when comparing approaches.}
\label{Procedure}
\end{algorithm}

\begin{table*}[h]
\centering
\caption{The average loglikelihoods of the 10 cross-validation splits along with the standard errors. For each split, the optimal regularization constants were obtained with a second 10-fold cross-validation. We used 5-fold cross-validation for the larger Kin8nm and Protein data sets. Bold values indicate a significant difference between equal and separate regularization at a 90$\%$ confidence level. Equal regularization never performs significantly better than separate regularization.}
\begin{adjustbox}{width=1\textwidth}
\begin{tabular}{l|ll|ll|ll}
  \multicolumn{7}{c}{\textbf{Loglikelihood}} \\
\toprule
 \textbf{Data Set} & \multicolumn{6}{c}{\textbf{Training Strategy}} \\
  & \multicolumn{2}{c|}{No Warm-up} & \multicolumn{2}{c|}{Warm-up} &\multicolumn{2}{c}{Warm-up fixed Mean} \\
 & \multicolumn{1}{c}{Equal} & \multicolumn{1}{c|}{Separate} & \multicolumn{1}{c}{Equal} & \multicolumn{1}{c|}{Separate} & \multicolumn{1}{c}{Equal} & \multicolumn{1}{c}{Separate} \\
& \multicolumn{1}{c}{regularization} & \multicolumn{1}{c|}{regularization} & \multicolumn{1}{c}{regularization} & \multicolumn{1}{c|}{regularization} & \multicolumn{1}{c}{regularization} & \multicolumn{1}{c}{regularization} \\
 \midrule
Boston Housing & $-2.61 \pm 0.105$ & $-2.61 \pm 0.105$ & $-2.59 \pm 0.0902$ & $-2.59 \pm 0.0902$ & $-2.67 \pm 0.0846$ & $-2.60 \pm 0.0683$ \\
Energy & $-1.25 \pm 0.227$ & $\bm{-0.685 \pm 0.100}$ & $-1.18 \pm 0.474$ & $-0.738 \pm 0.117$ & $-1.43 \pm 0.438$ & $-0.986 \pm 0.178$ \\
Yacht & $-0.599 \pm 0.147$ & $-0.482 \pm 0.141$ & $-0.249 \pm 0.121$ & $-0.216 \pm 0.208$ & $-0.972 \pm 0.161$ & $\bm{-0.519 \pm 0.109}$ \\
Concrete & $-3.37 \pm 0.0834$ & $-3.27 \pm 0.0891$ & $-3.23 \pm 0.153$ & $-3.23 \pm 0.153$ & $-3.30 \pm 0.0431$ & $\bm{-3.12 \pm 0.0651}$ \\
Wine quality red & $-0.994 \pm 0.0142$ & $-0.967 \pm 0.00659$ & $-0.960 \pm 0.0119$ & $-0.965 \pm 0.00494$ & $-0.957 \pm 0.0197$ & $-0.972 \pm 0.0148$ \\
Kin8nm & $1.27 \pm 0.0105$ & $\bm{1.28 \pm 0.0108}$ & $1.25 \pm 0.00691$ & $1.26 \pm 0.00870$ & $1.14 \pm 0.0147$ & $\bm{1.26 \pm 0.0108}$ \\
Protein & $-2.83 \pm 0.00958$ & $-2.84 \pm 0.0125$ & $-2.82 \pm 0.00861$ & $\bm{-2.80 \pm 0.00745}$ & $-2.85 \pm 0.0174$ & $-2.83 \pm 0.00921$ \\
\bottomrule
\end{tabular}
\end{adjustbox}
\label{LLresults}
\end{table*}

\begin{table*}[h]
\centering
\caption{The average RMSE-values of the 10 cross-validation splits along with the standard errors. For each split, the optimal regularization constants were obtained with a second 10-fold cross-validation. We used 5-fold cross-validation for the larger Kin8nm and Protein data sets. Bold values indicate a significant difference between equal and separate regularization at a 90$\%$ confidence level. Equal regularization never performs significantly better than separate regularization.}
\begin{adjustbox}{width=1\textwidth}
\begin{tabular}{l|ll|ll|ll}
  \multicolumn{7}{c}{\textbf{RMSE}} \\
\toprule
 \textbf{Data Set}& \multicolumn{6}{c}{\textbf{Training Strategy}} \\
  & \multicolumn{2}{c|}{No Warm-up} & \multicolumn{2}{c|}{Warm-up} &\multicolumn{2}{c}{Warm-up fixed Mean} \\
 & \multicolumn{1}{c}{Equal} & \multicolumn{1}{c|}{Separate} & \multicolumn{1}{c}{Equal} & \multicolumn{1}{c|}{Separate} & \multicolumn{1}{c}{Equal} & \multicolumn{1}{c}{Separate} \\
 & \multicolumn{1}{c}{regularization} & \multicolumn{1}{c|}{regularization} & \multicolumn{1}{c}{regularization} & \multicolumn{1}{c|}{regularization} & \multicolumn{1}{c}{regularization} & \multicolumn{1}{c}{regularization} \\
 \midrule
Boston Housing & $3.56 \pm 0.398$ & $3.56 \pm 0.398$ & $3.84 \pm 0.495$ & $3.84 \pm 0.495$ & $4.39 \pm 0.525$ & $3.75 \pm 0.399$ \\
Energy & $2.20 \pm 0.148$ & $\bm{0.468 \pm 0.0196}$ & $0.850 \pm 0.188$ & $\bm{0.507 \pm 0.0226}$ & $0.813 \pm 0.123$ & $0.604 \pm 0.0337$ \\
Yacht & $8.87 \pm 0.783$ & $\bm{3.31 \pm 0.697}$ & $1.00 \pm 0.179$ & $0.917 \pm 0.203$ & $1.12 \pm 0.127$ & $\bm{0.705 \pm 0.0839}$ \\
Concrete & $6.41 \pm 0.364$ & $6.57 \pm 0.247$ & $5.93 \pm 0.248$ & $5.83 \pm 0.182$ & $7.20 \pm 0.272$ & $\bm{5.54 \pm 0.222}$ \\
Wine quality red & $0.653 \pm 0.00924$ & $0.650 \pm 0.00941$ & $0.641 \pm 0.00635$ & $0.647 \pm 0.00931$ & $0.643 \pm 0.0102$ & $0.638 \pm 0.00774$ \\
Kin8nm & $0.0717 \pm 0.00107$ & $0.0706 \pm 0.00147$ & $0.0771 \pm 0.00104$ & $0.0757 \pm 0.00148$ & $0.0834 \pm 0.00130$ & $\bm{0.0701 \pm 0.000865}$ \\
Protein & $4.55 \pm 0.0249$ & $4.54 \pm 0.0106$ & $4.47 \pm 0.0146$ & $\bm{4.42 \pm 0.0246}$ & $4.59 \pm 0.0634$ & $4.49 \pm 0.0338$ \\
\bottomrule
\end{tabular}
\end{adjustbox}
\label{RMSEresults}
\vskip-0.15in
\end{table*}

\subsection{Architecture and training details} \label{architecture}
\begin{itemize}
	\item We use a split architecture, meaning that the network consists of two sub-networks that output a mean and a variance estimate. Each sub-network has two hidden layers with 40 and 20 hidden units and ELU \citep{clevert2015fast} activation functions. The mean-network has a linear transformation in the output layer and variance-network an exponential transformation to guarantee positivity. We also added a minimum value of $10^{-6}$ for numerical stability.
	\item All covariates and targets are standardized before training. 
	\item We use the Adam optimizer \citep{kingma2014adam} with gradient clipping set at value 5. We found that this greatly improves training stability in our experiments.
	\item We use a default batch-size of 32.
	\item We use 1000 epochs for each training stage. We found that this was sufficient for all networks to converge.
	\item We set the bias of the variance to 1 at initialization. This makes sure that the variance predictions are more or less constant at initialization.
\end{itemize}

\subsection{Results and discussion} \label{resultsanddiscussion}
\noindent The results are given in Tables \ref{LLresults} and \ref{RMSEresults}. Bold values indicate that for that specific training strategy (no warm-up, warm-up, or warm-up fixed mean) there is a significant difference between equal and separate regularization. This means that every row can have up to three bold values. Significance was determined by taking the differences per fold and testing if the mean of these differences is significantly different from zero using a two-tailed $t$-test at a 90$\%$ confidence level.

 We see that a warm-up is often very beneficial. For the yacht data set, we observe a considerable improvement in the RMSE when we use a warm-up period. A warm-up also drastically improves the result on the energy data set when we do not allow separate regularization. 

Generally, the difference between keeping the mean fixed after the warm-up and optimizing the mean and variance simultaneously after the warm-up is less pronounced. For a few data sets (Concrete, Kin8nm, Protein) we do observe a considerable difference in root-mean-squared error if we only consider equal regularization. If we allow separate regularization, however, these differences disappear.   

A separate regularization often drastically outperforms equal regularization. The energy data set gives the clearest example of this. For all three training strategies, a separate regularization performs much better than an equal regularization of the mean and variance. A similar pattern can be seen for the yacht data set. The optimal regularization for the variance was typically similar or an order of magnitude larger than the optimal regularization of the mean, never lower. We would like to stress that statistically significant results are difficult to obtain with only 5 to 10 folds but that the pattern emerges clearly: separate  regularization often improves the results while never leading to a significant decline.

Equal regularization and no warm-up perform as well as the other strategies for some data sets, although never considerably better. For Boston Housing, for example, using a warm-up and separate regularization yields very similar results as the other strategies. This can happen since the problem may be easy enough that the network is able to simultaneously estimate the mean and the variance without getting stuck. Additionally, while there is no reason to assume so a priori, the optimal regularization constant for the mean and the variance can be very similar. In fact, for the Boston Housing experiment we often found the same optimal regularization constant for the mean and variance during the cross-validation.

\section{Conclusion}
\noindent In this paper, we tested various training strategies for MVE networks. Specifically, we investigated whether following the recommendations of the original authors solves the recently reported convergence problems and we proposed a novel improvement, separate regularization. 

We conclude that the use of a warm-up period is often essential to achieving optimal results and fixes the convergence problems. Without this period, the network can fail to learn regions on which it performs poorly at the start of the training. We empirically observed that not using a warm-up period can lead to highly suboptimal results, both in terms of RMSE and loglikelihood.

We did not find evidence that clearly favors one of the strategies \textit{after} the warm-up, keeping the mean fixed or optimizing the mean and variance simultaneously. In theory, joint maximum likelihood estimation of the mean and the variance is advantageous. In practice, we did not observe significant differences. Optimizing the mean and variance simultaneously after the warm-up was seemingly only beneficial when separate regularization was not allowed. 

Current practice - see for instance the implementation of Concrete Dropout \citep{gal2017concrete} - is to enforce the same regularization constants for estimating the mean and variance, which implicitly assumes that both functions exhibit a similar degree of smoothness. However, our simulations indicate that this assumption is often violated. Real-world datasets often require a complex mean function that differs significantly from a constant function, while the variance function varies more smoothly, if at all. Therefore, it is more realistic to apply separate regularisation constants for the mean and variance functions. Our experiments demonstrate that this approach can indeed lead to significant improvements in model performance across various data sets.

\subsection{Recommendations}
\noindent Based on our experiments, we make the following recommendations when training an MVE network:
\begin{itemize}
	\item Use a warm-up. It is important to initialize the variance such that it is more or less constant for all inputs. Otherwise, some regions may be neglected. This is easily achieved by setting the bias of the variance neuron to 1 at initialization.
	\item Use gradient clipping. We found gradient clipping to yield more stable optimization when optimizing the mean and variance simultaneously. 
	\item Use separate regularization for the mean and variance. If a hyperparameter search is computationally infeasible, the variance should typically be regularized an order of magnitude stronger than the mean.
\end{itemize}

\subsection{Future work}
\noindent The results on separate regularization indicate that variance and mean functions are generally not equally smooth. It is therefore likely not optimal to use a similar architecture and training procedure for both. It would be interesting to investigate whether the use of a separate architecture and training procedure leads to further improvements.


\FloatBarrier
\section*{References}
\bibliographystyle{apalike}
\bibliography{../../references3}

\begin{thebibliography}{}

\bibitem[Abdar et~al., 2021]{abdar2021review}
Abdar, M., Pourpanah, F., Hussain, S., Rezazadegan, D., Liu, L., Ghavamzadeh, M., Fieguth, P., Cao, X., Khosravi, A., Acharya, U.~R., et~al. (2021).
\newblock A review of uncertainty quantification in deep learning: {{Techniques}}, applications and challenges.
\newblock {\em Information Fusion}.

\bibitem[Chaudhary et~al., 2022]{chaudhary2022flood}
Chaudhary, P., Leit{\~a}o, J.~P., Donauer, T., D'Aronco, S., Perraudin, N., Obozinski, G., {Perez-Cruz}, F., Schindler, K., Wegner, J.~D., and Russo, S. (2022).
\newblock Flood uncertainty estimation using deep ensembles.
\newblock {\em Water}, 14(19):2980.

\bibitem[Clevert et~al., 2015]{clevert2015fast}
Clevert, D.-A., Unterthiner, T., and Hochreiter, S. (2015).
\newblock Fast and accurate deep network learning by exponential linear units (elus).
\newblock {\em arXiv preprint arXiv:1511.07289}.

\bibitem[DeGroot, 1986]{degroot1986Probability}
DeGroot, M.~H. (1986).
\newblock {\em Probability and Statistics}.
\newblock {Addison-Wesley Pub. Co}, {Reading, Mass}, 2nd ed edition.

\bibitem[Dodge, 2008]{dodge2008concise}
Dodge, Y. (2008).
\newblock {\em The Concise Encyclopedia of Statistics}.
\newblock {Springer Science \& Business Media}.

\bibitem[Egele et~al., 2021]{egele2021autodeuq}
Egele, R., Maulik, R., Raghavan, K., Balaprakash, P., and Lusch, B. (2021).
\newblock {{AutoDEUQ}}: {{Automated}} deep ensemble with uncertainty quantification.
\newblock {\em arXiv preprint arXiv:2110.13511}.

\bibitem[Gal, 2016]{gal2016uncertainty}
Gal, Y. (2016).
\newblock {\em Uncertainty in Deep Learning}.
\newblock PhD thesis, University of Cambridge.

\bibitem[Gal and Ghahramani, 2016]{gal2016dropout}
Gal, Y. and Ghahramani, Z. (2016).
\newblock Dropout as a {{Bayesian}} approximation: {{Representing}} model uncertainty in deep learning.
\newblock In {\em International {{Conference}} on {{Machine Learning}}}, pages 1050--1059.

\bibitem[Gal et~al., 2017]{gal2017concrete}
Gal, Y., Hron, J., and Kendall, A. (2017).
\newblock Concrete dropout.
\newblock {\em Advances in Neural Information Processing Systems}, 30.

\bibitem[Gauss, 1823]{gauss1823theoria}
Gauss, C.-F. (1823).
\newblock {\em Theoria Combinationis Observationum Erroribus Minimis Obnoxiae}.
\newblock {Henricus Dieterich}.

\bibitem[{Hern{\'a}ndez-Lobato} and Adams, 2015]{hernandez2015probabilistic}
{Hern{\'a}ndez-Lobato}, J.~M. and Adams, R. (2015).
\newblock Probabilistic backpropagation for scalable learning of {{Bayesian}} neural networks.
\newblock In {\em International Conference on Machine Learning}, pages 1861--1869.

\bibitem[Heskes, 1997]{heskes1997practical}
Heskes, T. (1997).
\newblock Practical confidence and prediction intervals.
\newblock In {\em Advances in {{Neural Information Processing Systems}}}, pages 176--182.

\bibitem[H{\"u}llermeier and Waegeman, 2019]{hullermeier2019aleatoric}
H{\"u}llermeier, E. and Waegeman, W. (2019).
\newblock Aleatoric and epistemic uncertainty in machine learning: {{A}} tutorial introduction.
\newblock {\em arXiv preprint arXiv:1910.09457}.

\bibitem[Jain et~al., 2020]{jain2020maximizing}
Jain, S., Liu, G., Mueller, J., and Gifford, D. (2020).
\newblock Maximizing overall diversity for improved uncertainty estimates in deep ensembles.
\newblock In {\em Proceedings of the {{AAAI}} Conference on Artificial Intelligence}, volume~34, pages 4264--4271.

\bibitem[Khosravi and Nahavandi, 2014]{khosravi2014optimized}
Khosravi, A. and Nahavandi, S. (2014).
\newblock An optimized mean variance estimation method for uncertainty quantification of wind power forecasts.
\newblock {\em International Journal of Electrical Power \& Energy Systems}, 61:446--454.

\bibitem[Kingma and Ba, 2014]{kingma2014adam}
Kingma, D.~P. and Ba, J. (2014).
\newblock Adam: {{A}} method for stochastic optimization.
\newblock {\em arXiv preprint arXiv:1412.6980}.

\bibitem[Lakshminarayanan et~al., 2017]{lakshminarayanan2017simple}
Lakshminarayanan, B., Pritzel, A., and Blundell, C. (2017).
\newblock Simple and scalable predictive uncertainty estimation using deep ensembles.
\newblock In {\em Advances in {{Neural Information Processing Systems}}}, pages 6402--6413.

\bibitem[MacKay, 1992]{mackay1992practical}
MacKay, D.~J. (1992).
\newblock A practical {{Bayesian}} framework for backpropagation networks.
\newblock {\em Neural Computation}, 4(3):448--472.

\bibitem[Neal, 2012]{neal2012bayesian}
Neal, R.~M. (2012).
\newblock {\em Bayesian {{Learning}} for {{Neural Networks}}}, volume 118.
\newblock {Springer Science \& Business Media}.

\bibitem[Nix and Weigend, 1994]{nix1994estimating}
Nix, D.~A. and Weigend, A.~S. (1994).
\newblock Estimating the mean and variance of the target probability distribution.
\newblock In {\em Proceedings of 1994 {{IEEE International Conference}} on {{Neural Networks}} ({{ICNN}}'94)}, volume~1, pages 55--60. {IEEE}.

\bibitem[Pav, 2015]{pav2015moments}
Pav, S.~E. (2015).
\newblock Moments of the log non-central chi-square distribution.
\newblock {\em arXiv preprint arXiv:1503.06266}.

\bibitem[Pearce et~al., 2018]{pearce2018high}
Pearce, T., Brintrup, A., Zaki, M., and Neely, A. (2018).
\newblock High-quality prediction intervals for deep learning: {{A}} distribution-free, ensembled approach.
\newblock In {\em International Conference on Machine Learning}, pages 4075--4084.

\bibitem[Seitzer et~al., 2021]{seitzer2021pitfalls}
Seitzer, M., Tavakoli, A., Antic, D., and Martius, G. (2021).
\newblock On the pitfalls of heteroscedastic uncertainty estimation with probabilistic neural networks.
\newblock In {\em International Conference on Learning Representations}.

\bibitem[Shaikhina and Khovanova, 2017]{shaikhina2017handling}
Shaikhina, T. and Khovanova, N.~A. (2017).
\newblock Handling limited datasets with neural networks in medical applications: {{A}} small-data approach.
\newblock {\em Artificial intelligence in medicine}, 75:51--63.

\bibitem[Skafte et~al., 2019]{skafte2019reliable}
Skafte, N., J{\o}rgensen, M., and Hauberg, S. (2019).
\newblock Reliable training and estimation of variance networks.
\newblock {\em Advances in Neural Information Processing Systems}, 32.

\bibitem[Takahashi et~al., 2018]{takahashi2018student}
Takahashi, H., Iwata, T., Yamanaka, Y., Yamada, M., and Yagi, S. (2018).
\newblock Student-t variational autoencoder for robust density estimation.
\newblock In {\em {{IJCAI}}}, pages 2696--2702.

\bibitem[Theobald, 1974]{theobald1974generalizations}
Theobald, C.~M. (1974).
\newblock Generalizations of mean square error applied to ridge regression.
\newblock {\em Journal of the Royal Statistical Society: Series B (Methodological)}, 36(1):103--106.

\bibitem[{van Wieringen}, 2015]{van2015lecture}
{van Wieringen}, W.~N. (2015).
\newblock Lecture notes on ridge regression.
\newblock {\em arXiv preprint arXiv:1509.09169}.

\end{thebibliography}

\appendix
\onecolumn
\section{Details on the advantage of taking the variance into account} \label{advantages}
\subsection{General case} \label{generalcase}
\noindent As we stated in the main text, there are advantages to optimizing the full likelihood.  The resulting maximum-likelihood-estimate is consistent and asymptotically efficient \citep[chapter~7]{degroot1986Probability}. Specifically, no other consistent estimator can asymptotically have a lower variance.

This lower variance results in improved metrics such as RMSE. To see this, we analyze the expected squared error of a new observation: $\E{(y_{\text{new}} - \mu_{\hat{\theta}}(\bm{x}_{\text{new}}))^{2}}$. Here, we assume that there exists a true $\theta_{0}$ and that $\hat{\theta}$ is the maximum-likelihood-estimate. The expectation is taken over $y_{\text{new}}$ and the data with which $\hat{\theta}$ is created. The expected squared error is given by
\begin{align}
\nonumber &\E{(y_{\text{new}} - \mu_{\hat{\theta}}(\bm{x}_{\text{new}}))^{2}} \\
\nonumber & =	\E{(y_{\text{new}} - \mu_{\theta_{0}}(\bm{x}_{\text{new}}) + \mu_{\theta_{0}}(\bm{x}_{\text{new}}) - \mu_{\hat{\theta}}(\bm{x}_{\text{new}}))^{2}} \\
\nonumber &=\E{(y_{\text{new}} -\mu_{\theta_{0}}(\bm{x}_{\text{new}}))^{2}} + \E{(\mu_{\theta_{0}}(\bm{x}_{\text{new}}) - \mu_{\hat{\theta}}(\bm{x}_{\text{new}}))^{2}} \\
\nonumber &\approx\E{(y_{\text{new}} -\mu_{\theta_{0}}(\bm{x}_{\text{new}}))^{2}} + \E{(D_{\theta}\mu_{\theta_{0}}(\bm{x}_{\text{new}})(\hat{\theta} - \theta_{0}))^{2}} \\
\label{expectederror}& = C + \E{D_{\theta}\mu_{\theta_{0}}(\bm{x}_{\text{new}})(\hat{\theta} - \theta_{0})}^{2} + \V{D_{\theta}\mu_{\theta_{0}}(\bm{x}_{\text{new}})(\hat{\theta})}.
\end{align}
We now compare this to a different consistent estimator for $\theta_{0}$, denoted by $\tilde{\theta}$. By following the same derivation, we obtain an expected squared error of
\begin{equation}
\label{eq: expectederror2}
C + \E{D_{\theta}\mu_{\theta_{0}}(\bm{x}_{\text{new}})(\tilde{\theta} - \theta_{0})}^{2} + \V{D_{\theta}\mu_{\theta_{0}}(\bm{x}_{\text{new}})(\tilde{\theta})}.
\end{equation}
We can show that this is larger than (or equal to) the final line in equation \eqref{expectederror}. The first term, $C$, is equal for both \eqref{expectederror} and \eqref{eq: expectederror2}. Since $\hat{\theta}$ was obtained by using maximum likelihood optimization, we know two things. Firstly, we know that the second term in \eqref{expectederror} is asymptotically of lower order than the third term, and secondly, we know that
\[
\V{D_{\theta}\mu_{\theta_{0}}(\bm{x}_{\text{new}})(\tilde{\theta})} \geq \V{D_{\theta}\mu_{\theta_{0}}(\bm{x}_{\text{new}})(\hat{\theta})},
\]
which shows that our estimate $\tilde{\theta}$ asymptotically will have a larger expected quadratic error.

\subsection{Linear model} \label{linearmodeldetails1}
\noindent In the non-asymptotic regime, a similar result holds for a linear model. Taking the variance into account, leads to a lower-variance estimator. All derivations for the statements in this paper regarding linear models can be found in \citet{van2015lecture}.

We assume that we have a data set consisting of $n$ data points $(\bm{x}_{i}, y_{i})$, with $\bm{x}_{i} \in \mathbb{R}^{p}$ and $y \in \mathbb{R}$. With $X$, we denote the $n 
\times p$ design matrix which has the $n$ covariate vectors $\bm{x}_{i}$ as rows. With $Y$, we denote the $n\times 1$ vector containing the observations $y_{i}$. We assume $X$ to be of full rank and consider the linear model:
\begin{equation}
Y = X\beta + U, \quad U \sim \N{0}{\Sigma},
\label{eq: linearmodel}
\end{equation}
where $\Sigma$ can be any invertible covariance matrix, possibly heteroscedastic and including interaction terms.  Suppose this covariance matrix is known, then classical theory tells us that it is beneficial for our estimate of $\beta$ to take this into account. 

To see this, we will compare the linear model in Equation \eqref{eq: linearmodel} with a rescaled version that takes the covariance matrix into account. Since $\Sigma$ is positive semi-definite, we can write it as $BB^{T}$ and rescale our model by multiplying with $B^{-1}$:
\begin{align}
Z := B^{-1}Y &= B^{-1}X\beta + B^{-1}U \\
 &= \tilde{X}\beta + V, \quad V \sim \N{0}{I_{n}}
\end{align}
Both formulations lead to different estimators of $\beta$. The unscaled formulation leads to 
\[
\hat{\beta} = (X^{T}X)^{-1}X^{T}Y,
\]
and the second formulation leads to
\[
\hat{\beta}^{\ast} = (\tilde{X}^{T}\tilde{X})^{-1}\tilde{X}^{T}B^{-1}Y.
\]
Both estimators are linear unbiased estimators of $\beta$. However, the Gauss-Markov theorem \citep{gauss1823theoria} (see \citet{dodge2008concise} for a version that is not in Latin) tells us that the variance of $\hat{\beta}^{\ast}$ is lower than the variance of $\hat{\beta}$.
\newpage
\begin{namedtheorem}[Gauss-Markov]
	\citep{gauss1823theoria} \\
In the notation introduced above, consider the linear model $Y=X\beta + U$. Under the following assumptions:
\begin{enumerate}
	\item $\E{U}=\bm{0}$,
	\item $\V{U} = c I_{n}$,
	\item $X$ is of full rank,
\end{enumerate}
the ordinary least squares (OLS) estimator for $\beta$, $\hat{\beta}=(X^{T}X)^{-1}X^{T}Y$, has the lowest variance of all unbiased linear estimators of $\beta$, i.e., the difference of the covariance matrix of any unbiased linear estimator and the covariance matrix of the OLS estimator is positive semi-definite.
\end{namedtheorem}
We note that in the second formulation all the conditions of the  theorem are met. We therefore know that $\hat{\beta}^{\ast}$ has the lowest variance of all unbiased linear estimators of $\beta$ and thus in particular we know that it has a lower variance than $\hat{\beta}$. 

We want to emphasize that this leads to improved metrics such as RMSE. Let us for instance look at difference between the expected squared errors of a new pair $(\bm{x}_{\text{new}}, y_{\text{new}})$ when using $\hat{\beta}$ and $\hat{\beta}^{\ast}$:
\begin{align*}
& \E{(y_{\text{new}} - \bm{x}_{\text{new}}^{T}\hat{\beta})^{2} - (y_{\text{new}} - \bm{x}_{\text{new}}^{T}\hat{\beta}^{\ast})^{2}} \\
&\quad = \E{y_{\text{new}}^{2} - 2y_{\text{new}}\bm{x}_{\text{new}}^{T}\hat{\beta} + \bm{x}_{\text{new}}^{T}\hat{\beta}\hat{\beta}^{T}\bm{x}_{\text{new}}}  - \E{y_{\text{new}}^{2} - 2y_{\text{new}}\bm{x}_{\text{new}}^{T}\hat{\beta}^{\ast} + \bm{x}_{\text{new}}^{T}\hat{\beta}^{\ast}\hat{\beta}^{\ast T}\bm{x}_{\text{new}}} \\
&\quad = \E{\bm{x}_{\text{new}}^{T}\hat{\beta}\hat{\beta}^{T}\bm{x}_{\text{new}} - \bm{x}_{\text{new}}^{T}\hat{\beta}^{\ast}\hat{\beta}^{\ast T}\bm{x}_{\text{new}}} \\
&\quad = \bm{x}_{\text{new}}^{T}\left(\Sigma_{\hat{\beta}} - \Sigma_{\hat{\beta}^{\ast}}\right)\bm{x}_{\text{new}} \geq 0.
\end{align*}
We used that $\hat{\beta}$ and $\hat{\beta}^{\ast}$ are both unbiased and independent of $y_{\text{new}}$. In the final line, we applied the Gauss-Markov theorem that guarantees that $\Sigma_{\hat{\beta}} -\Sigma_{\hat{\beta}^{\ast}} $ is a positive semi-definite matrix.

\section{Details on the different optimal regularization constants for linear models} \label{linearmodeldetails2}
\noindent We consider two linear models that most closely resemble the scenario of an MVE network. The first model will estimate the mean while knowing the variance and the second model will estimate the log of the variance while knowing the mean. An MVE network often uses an exponential transformation in the output of the variance neuron to ensure positivity. The network then learns the log of the variance. We show that both models will generally have a different optimal regularization constant.
\subsection{Scenario 1: Estimating the mean with a known variance}

\noindent We use the same notation as in the previous example and assume a homoscedastic noise with variance $\sigma^{2}$. If we do not consider regularization, the goal is to find the estimator that minimizes the sum of squared errors,
\[
\sum_{i=1}^{n}(y_{i} - \bm{x}^{T}_{i}\beta)^{2}.
\]
The solution is given by
\[
\hat{\beta} = (X^{T}X)^{-1}X^{T}Y,
\]
for which we know that
\[
\E{\hat{\beta}} = \beta \quad \text{and} \quad  \V{\hat{\beta}} = \Sigma_{\hat{\beta}} = \sigma^{2}(X^{T}X)^{-1}.
\]
In particular, we used that  $\E{\epsilon}=0$ and $\V{\epsilon}=\sigma^{2}$. 

When we add a regularization constant, $\lambda$, the objective becomes to minimize
\[
\sum_{i=1}^{N}(y_{i} - \bm{x}^{T}_{i}\beta)^{2} + \lambda \sum_{j=1}^{p}\beta_{j}^{2}.
\]
The solution to this problem is given by
\[
\hat{\beta}(\lambda) = (X^TX + \lambda I_{p})^{-1}X^{T}Y.
\]
For simplicity, we assume that we have an orthonormal  basis, in which case $X^{T}X=I$. This does not change the essence of the argument and makes the upcoming comparison clearer. Our new estimate is no longer unbiased but has a lower variance:
\[
\E{\hat{\beta}(\lambda)} = \frac{1}{1 +\lambda} \beta \quad \text{en} \quad \V{\hat{\beta}(\lambda)} = \sigma^{2}(1+\lambda)^{-2}I_{p}
\]
Our goal is to answer the question what the optimal value for $\lambda$ is. We define optimal as the $\lambda$ for which

\[
\text{MSE}(\hat{\beta}(\lambda)):= \E{|\beta - \hat{\beta}(\lambda)|^{2}}
\]
is minimal.

\citet{theobald1974generalizations} showed that there exists $\lambda>0$ such that $\text{MSE}(\hat{\beta}(\lambda)) < \text{MSE}(\hat{\beta})$. Typically, the exact value of $\lambda$ is unknown, but in our controlled example, the optimal value can be derived analytically \citep{van2015lecture}: 
\[
\lambda^{\ast} = p \sigma^{2} (\beta^{T}\beta)^{-1}.
\]

\subsection{Scenario 2: Estimating the log variance with a known mean}
\noindent Next, we examine a linear model that estimates the logarithm of the variance. We again have $n$ datapoints $(\bm{x}_{i}, y_{i})$ where we assume the log of the variance to be a linear function of the covariates: \[
y_{i} = \mu_{i} + \epsilon, \quad \epsilon \sim \N{0}{e^{\bm{x}_{i}^{T}\tilde{\beta}}}
\]
We use the same covariates and for the targets we define:
\[
z_{i} := \log((y_{i} - \mu)^{2})) - C, \quad \text{with} \quad C = \psi(1/2) +\log(2),
\]
where $\psi$ is the digamma function. This somewhat technical choice for $C$ is made such that
\[
z_{i} = \log(\sigma^{2}(\bm{x}_{i})) + \tilde{\epsilon}, 
\]
where $\tilde{\epsilon}$ has expectation zero and a constant variance, as can be seen from the following derivation:
\begin{align*}
\log((y_{i} - \mu)^{2}) &= \log\left(\sigma^{2}(\bm{x}_{i}) \frac{(\bm{x}_{i} - \mu)^{2}}{\sigma^{2}(\bm{x}_{i})}\right)\\
&= \log(\sigma^{2}(\bm{x}_{i})) +\log\left(\frac{(\bm{x}_{i} - \mu)^{2}}{\sigma^{2}(\bm{x}_{i})} \right)\\
&= \log(\sigma^{2}(\bm{x}_{i})) +\log\left(\zeta \right), \; \zeta \sim \chi^{2}(1) \\
&= \log(\sigma^{2}(\bm{x}_{i})) + \tilde{\epsilon^{\ast}}
\end{align*}
The random variable $\tilde{\epsilon^{\ast}}$ has an expectation $C$ and a constant variance that does not depend on $\mu$ or $\sigma^{2}(\bm{x})$ \citep{pav2015moments}. The key result of this specific construction of $z$ is that we have recovered a linear model with additive noise that has zero mean and a constant variance.

This allows us to repeat the procedure from the previous subsection, i.e. minimizing the sum of squared errors with a regularization term. We obtain the following optimal regularization constant
\[
\tilde{\lambda}^{\ast} = p \V{\tilde{\epsilon}}(\tilde{\beta}^{T}\tilde{\beta})^{-1}.
\]
The conclusion is that for these two linear models, that most closely resemble the scenario of regularized neural networks that estimate the mean and log-variance, the optimal regularization constants rely on the true underlying parameters $\beta$ and $\tilde{\beta}$. Since, in general, these are different, a different regularization constant should be used.
\end{document}